\documentclass[lettersize,journal]{IEEEtran}

\usepackage{arydshln}
\usepackage{caption}
\usepackage{graphicx}
\usepackage{textcomp}
\usepackage{xcolor}
\usepackage{diagbox}
\usepackage{booktabs}
\usepackage{multirow}
\usepackage{bbding}
\usepackage{hyperref}
\usepackage{ulem}
\usepackage{array}
\usepackage{amsmath,amsfonts}
\usepackage{algorithmic}
\usepackage{algorithm}
\usepackage{array}
\usepackage[caption=false,font=normalsize,labelfont=sf,textfont=sf]{subfig}
\usepackage{textcomp}
\usepackage{stfloats}
\usepackage{url}
\usepackage{verbatim}
\usepackage{graphicx}
\usepackage{cite}
\begin{document}


\title{HSNet: Heterogeneous Subgraph Network for Single Image Super-resolution}

\author{Qiongyang Hu, Wenyang Liu, Wenbin Zou, Yuejiao Su, Lap-Pui Chau,~\IEEEmembership{Fellow,~IEEE,}
Yi Wang,~\IEEEmembership{Member,~IEEE} \\


\thanks{The research work was conducted in the JC STEM Lab of Machine Learning and Computer Vision funded by The Hong Kong Jockey Club Charities Trust. This research was partially funded by The Hong Kong Polytechnic University (PolyU) Start-up Fund for RAPs under the Strategic Hiring Scheme (P0047884).
}

\thanks{Qiongyang Hu, Wenbin Zou, Yuejiao Su, Lap-Pui Chau, and Yi Wang are with the Department of Electrical and
Electronic Engineering, The Hong Kong Polytechnic University, Hong Kong (e-mail: qiongyang.hu@connect.polyu.hk; alex14.zou@connect.polyu.hk; yuejiao.su@connect.polyu.hk; lap-pui.chau@polyu.edu.hk; yi-eie.wang@polyu.edu.hk).
}
\thanks{Wenyang Liu is with the School of Electrical and Electronic Engineering, Nanyang Technological University,
Singapore (wenyang001@e.ntu.edu.sg).}}



\maketitle

\begin{abstract}
Existing deep learning approaches for image super-resolution, particularly those based on CNNs and attention mechanisms, often suffer from structural inflexibility. Although graph-based methods offer greater representational adaptability, they are frequently impeded by excessive computational complexity. To overcome these limitations, this paper proposes the Heterogeneous Subgraph Network (HSNet), a novel framework that efficiently leverages graph modeling while maintaining computational feasibility. The core idea of HSNet is to decompose the global graph into manageable sub-components. First, we introduce the Constructive Subgraph Set Block (CSSB), which generates a diverse set of complementary subgraphs. Rather than relying on a single monolithic graph, CSSB captures heterogeneous characteristics of the image by modeling different relational patterns and feature interactions, producing a rich ensemble of both local and global graph structures. Subsequently, the Subgraph Aggregation Block (SAB) integrates the representations embedded across these subgraphs. Through adaptive weighting and fusion of multi-graph features, SAB constructs a comprehensive and discriminative representation that captures intricate interdependencies. Furthermore, a Node Sampling Strategy (NSS) is designed to selectively retain the most salient features, thereby enhancing accuracy while reducing computational overhead. Extensive experiments demonstrate that HSNet achieves state-of-the-art performance, effectively balancing reconstruction quality with computational efficiency. The code will be made publicly available.
\end{abstract}


\section{Introduction}
\IEEEPARstart{I}n the field of digital image processing, image upscaling is a fundamental and widely required operation. However, conventional interpolation-based methods often produce outputs with noticeable blurring and blocking artifacts, significantly compromising visual fidelity. To overcome these limitations, single image super-resolution (SISR) \cite{DBLP:journals/tbc/LinLYHC25,DBLP:journals/tbc/TangNLDD25,lepcha2023image, cong2023exploiting,park2021dynamic}  has emerged as a key technique that reconstructs high-resolution images from their low-resolution counterparts, thereby enhancing clarity and recovering fine-grained details. As a result, SISR has been extensively adopted in various domains, including medical imaging, remote sensing \cite{DBLP:conf/cvpr/ShermeyerE19}, and surveillance systems \cite{DBLP:conf/icmcs/YoshidaTDIM12}.

\begin{figure}[t]
    \centering
    \includegraphics[width=\linewidth]{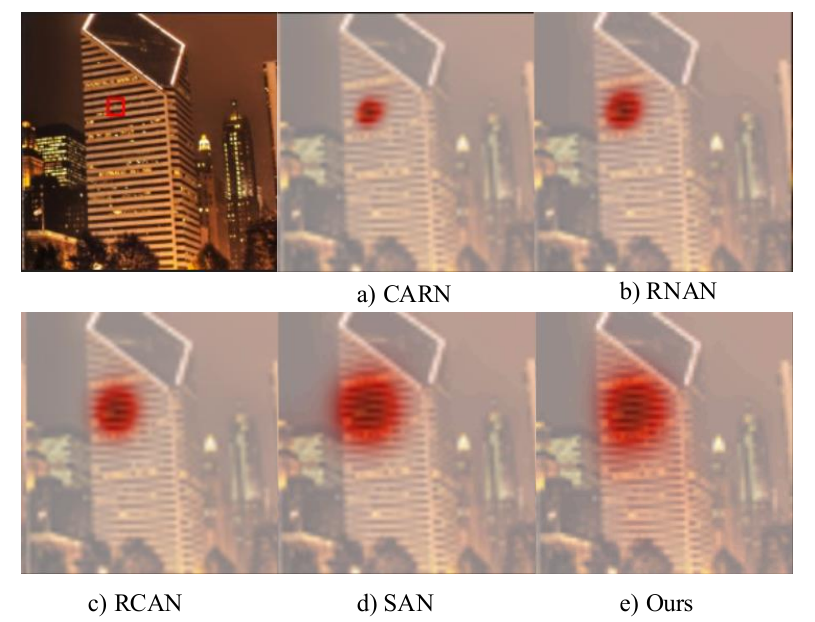}
    \caption{Different neural networks. 
    a) CARN \cite{ahn2018fast}: CNN-based network.
    b) RNAN \cite{zhang2019residual}: Non-local attention network.
    c) RCAN \cite{zhang2018image}: Channel attention network.
    d) SAN \cite{dai2019second}: Second order attention network.
    e) Ours: Graph-based network.
    }
    \label{fig:1}
 \vspace{-2mm}
\end{figure}

To further improve the quality of the reconstructed images, a wide range of advanced approaches have been developed. Among these, Convolutional Neural Networks (CNNs) \cite{kong2021classsr,lim2017enhanced,wang2017single} and window-based attention mechanisms \cite{zhang2024transcending,zhang2024hit} represent two dominant paradigms in the SISR field. As illustrated in Fig. \ref{fig:1}, CNN-based models such as CARN \cite{ahn2018fast} typically exhibit limited receptive fields, restricting their ability to capture long-range contextual information. In contrast, attention-based methods, exemplified by RNAN \cite{zhang2019residual}, RCAN \cite{zhang2018image}, and SAN \cite{dai2019second}, aim to overcome this limitation by adaptively focusing on salient regions, thereby enabling more effective modeling of global dependencies. However, this capability often comes at the cost of significantly increased computational complexity. More fundamentally, both CNN- and attention-based approaches share a common constraint: information aggregation is generally confined to predefined or locally constrained neighborhoods, limiting their adaptability in handling complex image structures and diverse reconstruction scenarios.

To address these challenges, we propose a novel Heterogeneous Subgraph Network (HSNet) that leverages graph structural properties to capture image features beyond predefined local neighborhoods. A major obstacle in applying graph-based methods to SISR lies in the high computational cost of modeling pairwise relationships across all pixels. To alleviate this issue while retaining the representational benefits of graphs, we introduce the Construct Subgraph Set Block (CSSB). Rather than building a single dense graph spanning the entire image, the CSSB efficiently generates a diverse set of subgraphs.
In this framework, a subgraph is defined as a subset of the global graph, consisting of selected vertices and their corresponding edges. By strategically constructing and processing multiple subgraphs, our approach preserves essential structural characteristics while enabling each subgraph to focus on specific local regions. This design not only substantially reduces computational complexity but also improves the flexibility and efficiency of feature extraction in HSNet.
Furthermore, to integrate the rich information captured across these subgraphs, we introduce the Subgraph Aggregation Block (SAB). The SAB is designed to extract and fuse features from the set of subgraphs, constructing a robust and heterogeneous global graph representation. Through this process, each node acquires a more comprehensive feature set, enabling the model to develop a holistic understanding of structural and relational patterns in the data. By incorporating multi-subgraph features, the SAB improves the model’s capacity to capture complex dependencies and subtle image details, thereby strengthening the predictive performance of HSNet.
As illustrated in Fig. \ref{fig:1}, our method exhibits a broader and more adaptive focus region compared to existing approaches.

In addition, to optimize the critical process of node selection, we introduce a strategy that effectively captures both fine-grained local details and high-level global structures from multiple perspectives.
Specifically, drawing inspiration from the Mamba scanning strategy \cite{zhu2024rethinking}, we design a Node Sampling Strategy (NSS) that identifies and retains the most important features that contribute to the learning objective, while efficiently reducing redundancy.
By selectively excluding less informative nodes, the proposed NSS preserves essential representational information, substantially lowers computational cost, and improves the overall efficacy of feature extraction. This results in enhanced accuracy and efficiency for the SISR task.

In addition, to optimize the critical process of node selection, we introduce a strategy that effectively captures both fine-grained local details and high-level global structures from multiple perspectives.
Specifically, drawing inspiration from the Mamba scanning strategy \cite{zhu2024rethinking}, we design a Node Sampling Strategy (NSS) that identifies and retains the most important features that contribute to the learning objective, while efficiently reducing redundancy.
By selectively excluding less informative nodes, the proposed NSS preserves essential representational information, substantially reduces computational cost, and improves the overall efficacy of feature extraction. This results in improved accuracy and efficiency for the SISR task, leading to improved model performance in the SISR task.
 The main contributions are shown as follows:

\begin{itemize}
  \item 
  We propose HSNet, a novel graph-structured framework that overcomes the rigid locality constraints of CNN- and attention-based models by representing image features through a diverse set of complementary subgraphs.
  \item 
  We introduce the Construct Subgraph Set Block (CSSB), which efficiently generates multiple subgraphs emphasizing different feature relationships, along with a Node Sampling Strategy (NSS) that retains salient features while reducing computational complexity.
\item 
We design the Subgraph Aggregation Block (SAB), which integrates information from individual subgraphs to form a comprehensive and discriminative feature representation, enhancing image reconstruction quality.
\item 
Extensive experiments demonstrate that HSNet achieves state-of-the-art performance across five SISR benchmarks. Ablation studies and visual analysis further validate the efficacy of the proposed components.
\end{itemize}

\section{Related work}
\subsection{Single Image Super-resolution}
Recent single image super-resolution (SISR) methods have shown remarkable advancements, incorporating innovative techniques and architectures to enhance image quality further.
Researchers have sought to refine CNN architectures \cite{dixit2024review,liu2023bitstream,liu2023bitstream1,DBLP:journals/tcsv/LiWWLLY25} to improve their performance.
Although CNNs excel in local regions, they struggle to effectively model long-range dependencies due to their limited receptive fields. To address this issue, researchers have proposed various techniques, such as using dilated convolutions \cite{zhang2018dcsr,liu2022multi} to increase the kernel size, thereby expanding the receptive field and enhancing SISR performance.

Another prominent approach is the use of Transformer models \cite{li2025srconvnet,li2025exploring,liu2025promptsr}, which effectively capture long-range dependencies due to the attention mechanism. However, this advantage is accompanied by considerable computational complexity.
To alleviate this computational burden, researchers have explored content-driven strategies \cite{hu2022fusformer,wei2024multi} that employ different information extraction methods for varying details. For instance, some studies \cite{ray2024cfat,xie2024dynamic} have attempted to establish connections between local windows using shifted windows to enhance interactions between them.

To further overcome the constraints of local windows, researchers have introduced category-based attention mechanisms \cite{zhang2024transcending} to establish long-range connections between similar structures within images. However, many models still remain bound by the limitations of local windows.
In contrast, our method uses subgraphs to capture long-range relationships rather than being confined to a single window, which can extract flexible representations.

\subsection{Graph Neural Network}
Some studies have begun to focus on irregular data structures, such as point clouds and social networks, utilizing graph-based methods to process these data more flexibly. Nevertheless, applying graph models directly to regular data—such as images—poses challenges that need to be addressed.
A graph is a universal data structure, and casting an image as a graph enhances flexibility and effectiveness in visual perception. This approach allows us to explore the complex relationships between pixels, more effectively handle irregularities, and improve the modeling of intricate patterns within the image. By utilizing graph-based methods \cite{yang2021image,tang2023graph,liu2022dual}, they can capture local and global dependencies better than traditional methods.
By treating image patches as nodes rather than individual pixels, Han et al. \cite{han2023vision} reduce complexity while effectively capturing spatial relationships and patterns.
Ref. \cite{ren2024key} treats patch features as graph nodes, using a Key-Graph Constructor to create a sparse, representative Key-Graph by selectively connecting essential nodes.
Tian et al. \cite{tian2024image} uses pixels instead of patches as nodes in the image graph. Our method constructs subgraphs based on the graph construction approach mentioned in \cite{zhou2020cross} to reduce the cost of building the entire graph. Then, we utilize the subgraph set to generate a robust Heterogeneous Graph.

\begin{figure*}[t]
    \centering   
    \includegraphics[width=\linewidth]{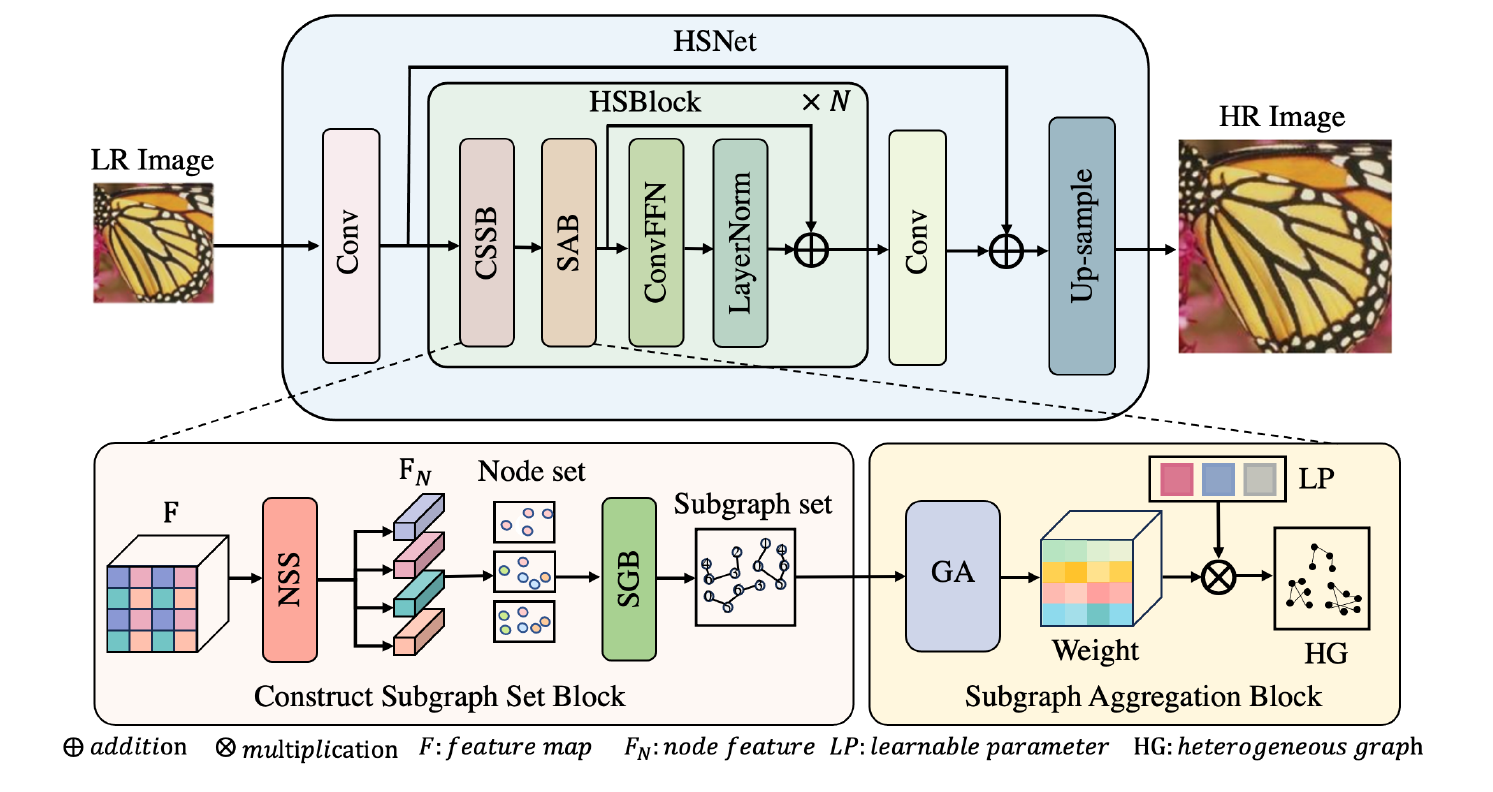}
    \caption{Architecture of HSNet: the core module is Heterogeneous Subgraph Block (HSBlock), which includes two modules: Construct Subgraph Set Block (CSSB) and Subgraph Aggregation Block (SAB). Among them, the CSSB consists of two components. Node Sampling Strategy (NSS) is responsible for sampling out nodes and forming a node set. Then the node set passes through the Subgraph Generation Block (SGB) to generate a subgraph set. Subgraph Aggregation Block (SAB) is responsible for generating a heterogeneous graph, which is composed of a Graph Aggregation (GA) block and a Learnable Parameter (LP).}
    \label{fig:2}

\end{figure*}

\subsection{Feature Scanning in Mamba}
Unlike CNNs and Transformers, the rise of Mamba \cite{zhang2024survey,liu2024vision,lei2024dvmsr} has opened up new research directions.
Mamba operates as a hardware-sensitive algorithm, utilizing recursive scanning and dynamic adaptability to maximize GPU capabilities while tackling various limitations inherent in conventional convolution methods.

Due to its inherent characteristics, Mamba processes image patches sequentially, offering a variety of scanning directions that enhance image understanding. In contrast to Vision Transformer \cite{liu2021swin}, which utilizes multi-head self-attention to analyze relationships between patches, Mamba sequentially processes them. This allows for multiple scanning directions of the available image patches.
Extensive research has been dedicated to investigating new scanning directions and their combinations to improve Mamba's capabilities in image understanding.
Various selective scan techniques utilized in Selective Scanning Methods are proposed for image and video processing. Many works \cite{liu2024vmamba,yu2025mambaout,chen2024rsmamba} have identified many common scanning directions, including ”Z”-shaped, diagonal, and ”S”-shaped patterns. Its diverse scanning methods accelerate the research process and enhance outcomes.
In addition, Pei et al. \cite{pei2025efficientvmamba} proposed an atrous-based selective scan approach through efficient skip sampling to leverage both global and local representational features. Inspired by the selective scan approach, we propose NSS to sample nodes that are essential for effective graph construction. 

\section{Our method}
To effectively apply graph structures directly to regular data, such as images, it is essential to transform the data into a graph representation, where pixels or regions of the image become nodes and relationships (such as adjacency or similarity) become edges. Each node should possess meaningful features that capture relevant information, such as color, texture, or spatial position, thereby enhancing the graph's utility in representing the image. By leveraging the flexibility of graphs, we can explore the relationships between irregularities within the data.

The graph structure is inherently more flexible and can transcend the limitations of windows to capture long-range information. However, its construction often requires a significant computational overhead. One direct solution is to reduce the number of nodes involved in the computation.
We propose the subgraph-based eterogeneous Subgraph Network (HSNet) to deal with this issue. 
In the following sections, we first present an overview of the overall model architecture. Then, we introduce the module of Construct Subgraph Set Block  (CSSB), which consists of two key components: Node Sampling Strategy (NSS) and Subgraph Generation Block (SGB). Subsequently, we describe the  Subgraph Aggregation Block (SAB), followed by a detailed explanation of the  Graph Aggregation (GA) component within SAB.


\subsection{Model Architecture}
The overall architecture of the model follows the traditional Super Resolution (SR) task structure and is divided into three parts, as shown in Fig.\ref{fig:2}. First, there is a Shallow Feature Extraction stage, typically implemented using a convolutional layer. 
The image is transformed into a feature map after passing through this layer. 
Next is the deep feature extraction phase, which is called Heterogeneous Subgraph Block (HSBlock).  The final component is the image reconstruction, which facilitates the increase in resolution.

Inspired by HetGNN \cite{zhang2019heterogeneous}, a representative method that achieves multi-source information integration through hierarchical aggregation, HSNet introduces the HSBlock—a dedicated module engineered to accommodate the diverse node and edge types in heterogeneous networks while facilitating the integration of multi-source information. The HSBlock is structurally composed of two core functional components: CSSB and SAB. Specifically, the CSSB undertakes the task of subgraph set generation: it extracts task-relevant information by adhering to specific criteria, whose configurations are dynamically adjusted based on the Node Sampling Strategy (NSS)——a method that encapsulates a suite of distinct node selection mechanisms. Then, a set of subgraphs is created by the SGB. 
Finally, the set of subgraphs is aggregated into a heterogeneous graph (HG) using the SAB.

\subsection{Construct Subgraph Set Block}
Spitz et al. \cite{spitz2018heterogeneous} point out that traditional homogeneous subgraph features overlook the multi-type node/edge information of heterogeneous networks, thus necessitating the targeted extraction of topological and semantic features from heterogeneous subgraphs.
The CSSB module is designed for this core need: it covers multi-level network structures via cross-scale sampling, captures node correlations with similarity calculations, retains node elements like pixel blocks while incorporating multi-type, multi-dimensional edges, laying a foundation for feature presentation and mining as a key link between heterogeneous info deficiency and feature extraction goals.
The CSSB begins by extracting subgraphs from the large graph using specific strategies.
We employ NSS and SGB to construct different subgraphs.
Each subgraph not only represents different subspaces but also dynamically selects neighboring nodes relevant to the target node for efficient information aggregation. By implementing different sampling strategies (such as local and global sampling), the model captures information at various scales. Local sampling focuses on relationships between nearby nodes, while global sampling captures structural features of the entire graph, thereby improving the model's understanding of complex graph data.

\subsubsection{Node Sampling Strategy}

We propose the NSS, a selective sampling approach to identify and preserve the most informative features, enabling more flexible feature selection. 
For the feature map $F \in R^{H \times W \times C}$, we employ a stride of 2 for selection, dividing the feature space into four distinct parts. The height (H) and width (W) are each segmented into four regions, which can be expressed with the following formula:
\begin{equation}
\begin{aligned}
F_{1} &= \{ F_{i,j,k,l} \in F \mid i, j \in \mathbb{Z}, \, k \equiv 0 \, (\text{mod} \, 2), \, l \equiv 0 \, (\text{mod} \, 2) \}, \\
F_{2} &= \{ F_{j,i,k,l} \in F \mid i, j \in \mathbb{Z}, \, k \equiv 0 \, (\text{mod} \, 2), \, l \equiv 1 \, (\text{mod} \, 2) \}, \\
F_{3} &= \{ F_{i,j,k,l} \in F \mid i, j \in \mathbb{Z}, \, k \equiv 0 \, (\text{mod} \, 2), \, l \equiv 1 \, (\text{mod} \, 2) \}, \\
F_{4} &= \{ F_{j,i,k,l} \in F \mid i, j \in \mathbb{Z}, \, k \equiv 1 \, (\text{mod} \, 2), \, l \equiv 1 \, (\text{mod} \, 2) \}.
\end{aligned}
\end{equation}
In this equation, \( F_{1} \) represents the first subset, consisting of elements \( F_{i,j,k,l} \in F \) where \( i, j \in \mathbb{Z} \) (indicating that \( i \) and \( j \) are integers), and both indices \( k \) and \( l \) are even, as expressed by \( k \equiv 0 \, (\text{mod} \, 2) \) and \( l \equiv 0 \, (\text{mod} \, 2) \). The subset \( F_{2} \) includes the transposed elements \( F_{j,i,k,l} \in F \), where \( k \) is even and \( l \) is odd, denoted by \( l \equiv 1 \, (\text{mod} \, 2) \). The subset \( F_{3} \) consists of elements \( F_{i,j,k,l} \in F \) with \( k \) even and \( l \) odd, while \( F_{4} \) includes the transposed elements \( F_{j,i,k,l} \) where both \( k \) and \( l \) are odd, represented by \( k \equiv 1 \, (\text{mod} \, 2) \) and \( l \equiv 1 \, (\text{mod} \, 2) \). 

After processing the selected features, resulting in dimensions of $F_{i} \in R^{H/p,W/p,C}$, where $p$ is the number of subsets. This effectively samples $F$ in four different ways, allowing us to capture data features from various perspectives, thereby enriching the feature representation. Subsequently, we recombine them to reconstruct the feature map $F_{concat} \in R^{H,W,C}$, as shown in the following formula:
\begin{equation}
F_{concat} = [F_1,F_2,F_3,F_4].
\end{equation}
This reconstruction enables the model to capture contextual information more broadly, enhancing its understanding.

\begin{figure}[t]
    \centering
     \includegraphics[width=\linewidth]{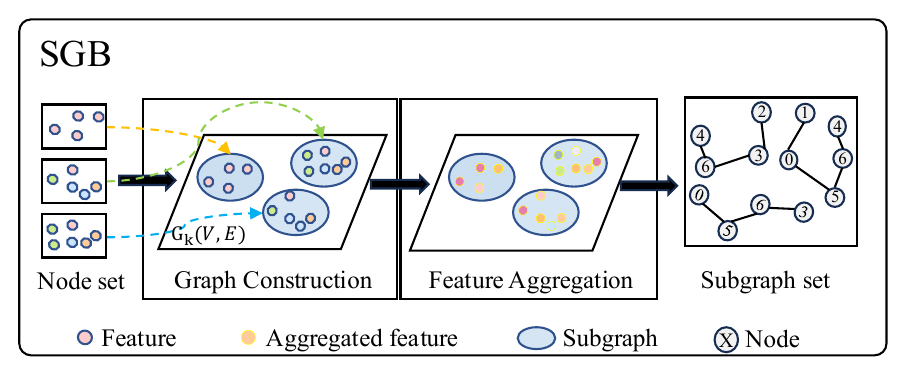}
    \caption{Illustration of Subgraph Generation Block (SGB).
    }
    \label{fig:3}
\end{figure}
\subsubsection{Subgraph Generation Block}
The SGB's role is to create subgraphs, as illustrated in the Fig.\ref{fig:3}. The SGB builds each subgraph by choosing specific nodes and edges.
Several studies have additionally employed prior knowledge of image self-similarity to enhance super-resolution (SR) outcomes. Zhou et al. \cite{zhou2020cross} proposed a cross-scale internal graph neural network (IGNN) that incorporates a graph neural network (GNN) framework to account for textural features across various scales. Specifically, the GNN's aggregation of self-similar cross-scale feature patches provides an insight, i.e., by integrating information with similar features across different scales, we can achieve a more comprehensive understanding and extraction of deep-level features in images. 

\textbf{Graph Construction.}
To obtain cross-scale features, we first use two different sampling methods to select patches at different scales. Grid sampling is employed, where patches are generated at fixed intervals at each scale. This method ensures uniform coverage of feature areas at each scale and helps capture detailed structural information. Then, we utilize the Structural Similarity Index Metric (SSIM) to evaluate the similarity between image patches. Moreover, for each patch, we compute its Euclidean distance to all other patches, retaining the $K$ edges with the smallest distances. This process culminates in the construction of a graph $G(V,E)$, where $V$ represents the set of vertices corresponding to each patch, and $E$ denotes the edges that connect these vertices. The weights of these edges serve as quantitative measures of similarity between the connected nodes, thereby facilitating a nuanced understanding of the relationships among the patches.
Therefore, similar feature patches and their interconnections are modeled as nodes and edges, respectively. This graph structure enables the aggregation of similar feature patches while preserving the spatial relationships among the nodes.

\textbf{Feature Aggregation.} By aggregating similar features, the characteristic information of adjacent nodes is integrated, providing a more comprehensive understanding of the context. The aggregation operation can be formulated as follows:

\begin{equation}
y_i = 1/C(x) \sum f(x_i,x_j)g(x_j),
\end{equation}
where $x_i$ denote the patch located at position $i$ and $x_j$ represent the patch at position $j$. The function $g(x_j)$ encapsulates the feature value of the patch $x_j$, while the function $f(x_i,x_j)$ is employed to compute the aggregation weights for $g(x_j)$. To ensure the effectiveness of the aggregation process, we define a normalized factor $C(X)$, which is calculated as the sum of the aggregation weights across all relevant patches. This normalization step is crucial as it guarantees that the aggregation weights are proportionate and maintain a consistent scale, allowing for a more accurate and effective combination of features from similar patches. This intricate approach not only enhances the feature extraction process but also contributes to a robust representation of the underlying image data.

\subsection{Subgraph Aggregation Block}
HSNet decomposes the global graph into multiple subgraphs, with each subgraph focusing on specific local semantics. These subgraphs are then fused via the Subgraph Aggregation Block (SAB). In essence, it alleviates semantic confusion through the "subgraph decomposition-aggregation" process, which is consistent with the core requirement of local semantic preservation in Heterogeneous Path-based Network (HPN) \cite{ji2021heterogeneous}.
HPN’s semantic fusion mechanism achieves multi-semantic integration by learning the importance of meta-paths, while the SAB of HSNet realizes weighted fusion of subgraph features through Learnable Parameters (LP). Both approaches balance the contribution of heterogeneous semantics by leveraging dynamic weights.
The Subgraph Aggregation Block (SAB) is introduced to consolidate the diverse features encoded by each subgraph. Through an adaptive process of weighting and combining information across various graph perspectives, SAB synthesizes a holistic feature representation, effectively modeling intricate interrelations among the data. Specifically, after constructing the subgraph set, we use the Graph Aggregation (GA) block, a multi-head attention mechanism to capture intrinsic relationships within the subgraphs and learn their feature representations. Each attention head independently learns the attention weights and feature representations for different subgraphs, allowing it to capture various types of relationships and features. This diversity enhances the model's expressive capability, with some heads focusing on local features (such as relationships among neighboring nodes) while others capture global features (such as the structure of the entire graph). This multi-level feature learning significantly improves the model's overall performance.



\begin{figure}[t]
    \centering
    \includegraphics[width=\linewidth]{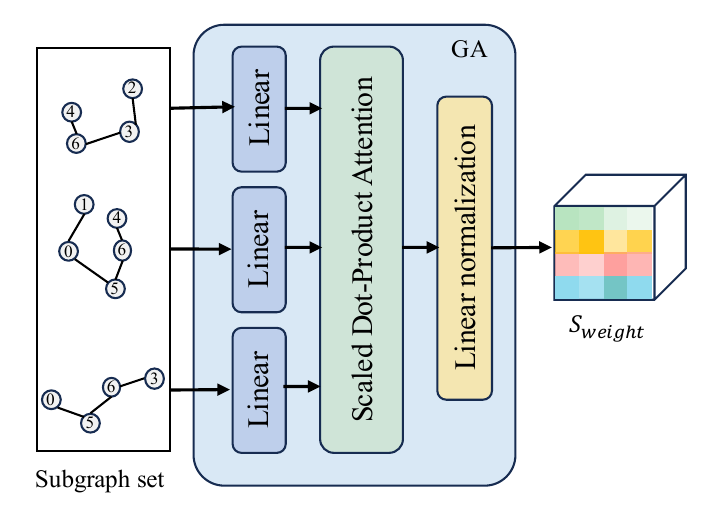}
    \caption{Illustration of Graph Aggregation (GA) block.
    }
    \label{fig:4}
\end{figure}

\subsubsection{Graph Aggregation}
The attention module is designed to deeply explore intrinsic relationships within subgraphs and learn their feature representations, as shown in Fig.\ref{fig:4}. Operating through multiple attention heads in parallel, it allows the model to analyze data from various perspectives. Each attention head independently calculates attention weights and feature representations, focusing on different aspects, such as local relationships between neighboring nodes or the global structure of the graph. This diversity enhances the model's expressive capability by capturing a wide range of relationships and features.

The input consists of several sets of feature representations, typically encoded as vectors. These can represent various aspects of the data, such as nodes in a graph.
Let $S^{(k)}$ represent the feature vectors for the  $k^{th}$ subgraph.
\begin{equation}
S^{(k)} = \{s_1^{(k)}, s_2^{(k)}, \ldots, s_n^{(k)}\},
\end{equation}
where n is the number of nodes in the subgraph.

Each input set undergoes a linear transformation through separate linear layers. This transforms the original feature vectors into three different representations: queries $(Q)$, keys $(K)$, and values $(V)$. 

\begin{equation}
Q^{(k)} = W_Q^{(k)} S^{(k)}, \quad K^{(k)} = W_K^{(k)} S^{(k)}, \quad V^{(k)} = W_V^{(k)} S^{(k)},
\end{equation}
where $W_Q^{(k)}$, $W_K^{(k)}$,$W_V^{(k)}$ are learnable weight matrices.
Then, we perform attention operations on different vectors, denoting them as $A^{(k)} = Attention(W_Q^{(k)}, W_K^{(k)},W_V^{(k)})$.

The feature representations are updated by aggregating the values based on the attention weights, as shown in the following equation:
\begin{equation}
S'^{(k)} = A^{(k)} V^{(k)}.
\end{equation}

The weights generated from the subset graph set after applying GA can be expressed as:
\begin{equation}
S_{weight} = \{S'^{(1)}, S'^{(2)}, \ldots, S'^{(J)}\},
\end{equation}
where $J$ represents the total number of subgraphs within the subset.

\subsubsection{Combination}
The model organically combines different subgraphs using a set of learnable dynamic parameters. These parameters can automatically adjust based on feedback during the training process, enhancing the model's adaptability. Depending on the types of relationships, the model can apply weighted fusion based on their importance, ensuring that more significant relationships have a greater impact on the final results, thus improving the model's predictive capability and interpretability.
Through the above structure, we enhance the representation within each node in the Heterogeneous Graph. 
The outputs from GA are combined using the dynamic Learnable Parameter (LP) $\alpha_n $.
\begin{equation}
HG = \alpha_1 S'^{(1)} + \alpha_2 S'^{(2)} + \ldots + \alpha_J S'^{(J)}
\end{equation}
where J represents the total number of learnable parameters.




\section{ EXPERIMENT}
\subsection{Experimental details}
\textbf{Datasets.} 
We adopt the standard training framework from recent super-resolution (SR) studies to ensure a fair evaluation. Our training dataset uses DIV2K for lightweight model training. 
We compare the performance of our model against several SR baselines, evaluated on the Set5 \cite{bevilacqua2012low}, Set14 \cite{zeyde2010single}, BSDS100 \cite{martind2001adatabaseof}, Urban100 \cite{huang2015single}, and Manga109 \cite{matsui2017sketch} datasets.

\begin{table*}[tb]
\caption{Comparison of HSNet with recent SR methods. Best and second best results are colored with \textcolor{red}{red} and \textcolor{blue}{blue}. }
\centering
\setlength{\tabcolsep}{1mm}
\begin{tabular}{cccccccccccccc}
\toprule[1pt]
{Method} & Scale & Params & \multicolumn{2}{c}{SET5}& \multicolumn{2}{c}{SET14}& \multicolumn{2}{c}{B100} & \multicolumn{2}{c}{Urban100} & \multicolumn{2}{c}{Manga109} \\
{} & && $PSNR$ & $SSIM$ & $PSNR$ & $SSIM$& $PSNR$ & $SSIM$& $PSNR$ & $SSIM$& $PSNR$ & $SSIM$\\
\midrule
CARN (2018) \cite{ahn2018fast} & $\times2$ & 1592K  &37.76 & 0.9590 & 33.52 & 0.9166 & 32.09 & 0.8978 & 31.92 & 0.9256 & 38.36 & 0.9765\\
IMDN (2019) \cite{hui2019lightweight} & $\times2$  & 694K   & 38.00 & 0.9605 & 33.63 & 0.9177 & 32.19 & 0.8996 & 32.17 & 0.9283 & 38.88 & 0.9774 \\
LAPAR-A(2020) \cite{li2020lapar} & $\times2$  & 548K &38.01 & 0.9605 & 33.62 & 0.9183 & 32.19 & 0.8999 & 32.10 & 0.9283 & 38.67 & 0.9772\\


SwinIR-light(2021) \cite{liang2021swinir} & $\times2$  & 878K & 38.14 & 0.9611 & 33.86 & 0.9206 & 32.31 & 0.9012 & 32.76 & 0.9340 & 39.12 & 0.9783 \\

ELAN-light (2022) \cite{zhang2022efficient} & $\times2$  & 582K & 38.17 & 0.9611 & 33.94 & 0.9207 & 32.30 & 0.9012 & 32.76 & 0.9340 & 39.11 & 0.9782\\

SRFormer-light (2023) \cite{zhou2023srformer} & $\times2$  & 853K &38.23 & 0.9613 & 33.94 & 0.9209 & 32.36 & 0.9019 & 32.91 & 0.9353 & 39.28 & 0.9785 \\
SAFMN (2023) \cite{sun2023spatially} & $\times2$  & 228K & 38.00 & 0.9605 & 33.54 & 0.9177 &  32.16 & 0.8995 & 31.84 & 0.9256 & 38.71 & 0.9771\\

MambaIR-light (2024) \cite{guo2024mambair} & $\times2$  & 530K	& 38.15	& 0.9610 &	33.84	& 0.9207 &	32.31 &	0.9013 &	32.86	& 0.9343	& 39.35	& 0.9786 
\\
FIWHN (2024) \cite{li2024efficient} & $\times2$ & 705K & 38.16 & 0.9613 & 33.73 & 0.9194 & 32.27 & 0.9007 & 32.75 & 0.9337 & 39.07& 0.9782 \\
SeemoRe-T(2024) \cite{zamfir2024see}  & $\times2$  & 220K &38.06 & 0.9608 & 33.65 & 0.9186 & 32.23 & 0.9004 & 32.22 & 0.9286 & 39.01 & 0.9777\\

IPG-tiny (2024) \cite{tian2024image} & $\times 2$ & 872K & 38.27 & 0.9616 &  \textcolor{red}{34.24} &  \textcolor{blue}{0.9236} & 32.35 & 0.9018 & 33.04 & 0.9359 & 39.31 & 0.9786 \\ 
CATANet (2025) \cite{liu2025catanet} & $\times2$  & 477K  & \textcolor{blue}{38.28} & \textcolor{blue}{0.9617} & 33.99 &  0.9217 & \textcolor{blue}{32.37} &  \textcolor{blue}{0.9023} &  \textcolor{blue}{33.09} & \textcolor{blue}{0.9372} &  \textcolor{blue}{39.37} &  \textcolor{blue}{0.9784}\\

GMN(2025) \cite{gendy2025lightweight} & $\times2$  & 1110k	& 38.16	& 0.9611 &	33.88 &	0.9201	& 32.30	& 0.9012	& 32.55 &	0.9325	& 39.17	& 0.9781 \\

\textbf{HSNet} & $\times2$  & 870K  & \textcolor{red}{38.29} & \textcolor{red}{0.9620} & \textcolor{blue}{34.22} & \textcolor{red}{0.9241} & \textcolor{red}{32.38} & \textcolor{red}{0.9031} & \textcolor{red}{33.27} & \textcolor{red}{0.9391} & \textcolor{red}{39.38} & \textcolor{red}{0.9802}\\

\midrule
CARN (2018) \cite{ahn2018fast} & $\times3$  & 1592K & 34.29 & 0.9255 & 30.29 & 0.8407 & 29.06 & 0.8034 & 28.06 & 0.8493 & 33.43 & 0.9427\\
IMDN (2019) \cite{hui2019lightweight}  & $\times3$ & 703K & 34.36 & 0.9270 & 30.32 & 0.8417 & 29.09 & 0.8046 & 28.17 & 0.8519 & 33.61 & 0.9445\\
LAPAR-A(2020) \cite{li2020lapar} & $\times3$ & 594K &  34.36 & 0.9267 & 30.34 & 0.8421 & 29.11 & 0.8054 & 28.15 & 0.8523 & 33.51 & 0.9441\\


SwinIR-light(2021) \cite{liang2021swinir}  & $\times3$ & 886K &  34.62 & 0.9289 & 30.54 & 0.8463 & 29.20 & 0.8082 & 28.66 & 0.8624 & 33.98 & 0.9478 \\
ELAN-light (2022) \cite{zhang2022efficient} &$\times3$ & 590K & 34.64 & 0.9288 & 30.55 & 0.8463 & 29.21 & 0.8081 & 28.69 & 0.8624 & 34.00 & 0.9478\\
SwinIR-light(2021) \cite{liang2021swinir}  & $\times3$ & 886K &  34.62 & 0.9289 & 30.54 & 0.8463 & 29.20 & 0.8082 & 28.66 & 0.8624 & 33.98 & 0.9478
\\
SRFormer-light (2023) \cite{zhou2023srformer} & $\times3$& 861K &  34.67 & 0.9296 & 30.57 & 0.8469 & 29.26 & 0.8099 & 28.81 & 0.8655 & 34.19 & 0.9489\\
SAFMN (2023) \cite{sun2023spatially} & $\times3$ & 233K & 34.34 & 0.9267 & 30.33 & 0.8418 & 29.08 & 0.8048 & 27.95 & 0.8474 & 33.52 & 0.9437\\

MambaIR-light (2024) \cite{guo2024mambair}  &$\times3$ & 538K	&	34.62 &	0.9286	& 30.54	& 0.8459	& 29.23	& 0.8083	& 28.73 &	0.8635	& 34.26 &	0.9482 \\
FIWHN (2024) \cite{li2024efficient}  & $\times3$ & 713K & 34.50 & 0.9283 & 30.50 & 0.8451 & 29.19 & 0.8077 & 28.62 & 0.8607 & 33.97& 0.9472 \\
SeemoRe-T(2024) \cite{zamfir2024see}   & $\times3$ & 225K & 34.46 & 0.9276 & 30.44 & 0.8445 & 29.15 & 0.8063 & 28.27 & 0.8538 & 33.92 & 0.9460\\
IPG-tiny (2024) \cite{tian2024image}  & $\times3$ & 878K & 34.64 & 0.9292 & 30.61 & 0.8470 & 29.26 & 0.8097 & 28.93 & 0.8666 & 34.30 & 0.9493\\
CATANet (2025) \cite{liu2025catanet} & $\times3$ & 550K & \textcolor{blue}{34.75} & \textcolor{blue}{0.9300} & \textcolor{blue}{30.67} & \textcolor{blue}{0.8481} & \textcolor{blue}{29.28} & \textcolor{blue}{0.8101} &  \textcolor{blue}{29.04} &  \textcolor{blue}{0.8689} &  \textcolor{blue}{34.40} &  \textcolor{blue}{0.9499} \\

GMN(2025) \cite{gendy2025lightweight} &	$\times3$ & 1120K	&	34.60 &	0.9291 &	30.37	& 0.8454	& 29.23	& 0.8091 & 	28.56	& 0.8609	& 34.14	& 0.9482 \\

\textbf{HSNet} & $\times3$ & 875K & \textcolor{red}{34.76} & \textcolor{red}{0.9301} & \textcolor{red}{30.69} & \textcolor{red}{0.8490} & \textcolor{red}{29.30} & \textcolor{red}{0.8102} & \textcolor{red}{29.07} & \textcolor{red}{0.8691} & \textcolor{red}{34.42} & \textcolor{red}{0.9512} \\

\midrule
CARN (2018) \cite{ahn2018fast}  & $\times4$ & 1592K &  32.13 & 0.8937 & 28.60 & 0.7806 & 27.58 & 0.7349 & 26.07 & 0.7837 & 30.42 & 0.9070\\
IMDN (2019) \cite{hui2019lightweight} &$\times4$  & 715K & 32.21 & 0.8948 & 28.58 & 0.7811 & 27.56 & 0.7353 & 26.04 & 0.7838 & 30.45 & 0.9075\\
LAPAR-A(2020) \cite{li2020lapar}  & $\times4$  &  659K & 32.15 & 0.8944 & 28.61 & 0.7818 & 27.61 & 0.7366 & 26.14 & 0.7871 & 30.42 & 0.9074\\

SwinIR-light(2021) \cite{liang2021swinir} &$\times4$  & 897K & 32.44 & 0.8976 & 28.77 & 0.7858 & 27.69 & 0.7406 & 26.47 & 0.7980 & 30.92 & 0.9151\\

ELAN-light (2022) \cite{zhang2022efficient} & $\times4$  & 601K & 32.43 & 0.8975 & 28.78 & 0.7858 & 27.69 & 0.7406 & 26.54 & 0.7982 & 30.92 & 0.9150\\
SwinIR-light(2021) \cite{liang2021swinir}  & $\times4$ & 897K  & 32.44 & 0.8976 & 28.77 & 0.7858 & 27.69 & 0.7406 & 26.47 & 0.7980 & 30.92 & 0.9151\\
SRFormer-light (2023) \cite{zhou2023srformer}   & $\times4$  & 873K & 32.51 & 0.8988 & 28.82 & 0.7872 & 27.73 & 0.7422 & 26.67 & 0.8032 & 31.17 & 0.9165\\
SAFMN (2023) \cite{sun2023spatially} & $\times4$  & 240K & 32.18 & 0.8948 & 28.60 & 0.7813 & 27.58 & 0.7359 & 25.97 & 0.7809 & 30.43 & 0.9063\\

MambaIR-light (2024) \cite{guo2024mambair}  & $\times4$ 	& 549K	&	32.41	& 0.8974	& 28.76	& 0.7849 &	27.70	& 0.7400 & 	26.53 &	0.7983 &	31.05	& 0.9144\\
FIWHN (2024) \cite{li2024efficient} & $\times4$ & 725K & 32.30 & 0.8967 & 28.76 & 0.7849 & 27.68 & 0.7400 & 26.57 & 0.7989 & 30.93 & 0.9131 \\
SeemoRe-T(2024) \cite{zamfir2024see}  & $\times4$ 	& 232K & 32.31 & 0.8965 & 28.72 & 0.7840 & 27.65 & 0.7384 & 26.23 & 0.7883 & 30.82 & 0.9107\\
IPG-tiny (2024) \cite{tian2024image}  & $\times4$  & 887K & 32.51 & 0.8987 & 28.85 & 0.7873 & 27.73 & 0.7418 & 26.78 & 0.8050 & 31.22 & 0.9176\\

CATANet (2025) \cite{liu2025catanet}  & $\times4$  & 535K &  \textcolor{blue}{32.58} &  \textcolor{blue}{0.8998} &  \textcolor{blue}{28.90} & \textcolor{blue}{0.7880} &  \textcolor{blue}{27.75} &  \textcolor{blue}{0.7427} &  \textcolor{blue}{26.87} &   \textcolor{blue}{0.8081} &  \textcolor{blue}{ 31.31} &  \textcolor{red}{0.9183}\\

GMN(2025) \cite{gendy2025lightweight} & $\times4$ 	& 1134K	&	32.40	& 0.8979 &	28.66	& 0.7868	& 27.70	& 0.7416 &	26.40 &	0.7963	& 30.98	& 0.9143 \\

\textbf{HSNet}  & $\times4$  & 882K & \textcolor{red}{32.60} & \textcolor{red}{0.9012} & \textcolor{red}{28.93} & \textcolor{red}{0.7886} & \textcolor{red}{27.78} & \textcolor{red}{0.7437} & \textcolor{red}{26.89} & \textcolor{red}{0.8089} &  \textcolor{red}{31.33} & \textcolor{blue}{0.9181}  \\

\bottomrule
\end{tabular}

\label{tab:1} 
\end{table*}

\begin{table}[t]
\caption{Param and Flops comparison of existing lightweight SR models on Urban100 ($\times4$) and on Manga109 ($\times4$). The best results are highlighted in \textbf{bold}. `C' represents CNN-based methods, `T' represents Transformer-based methods, and `G' represents graph-based methods.}

\resizebox{\linewidth}{!}{
\centering

\begin{tabular}{ccccccc}
\toprule[1pt]
Method & Type  & Param & Flops &  Urban100 & Manga109\\
\midrule
CARN \cite{ahn2018fast} & C & 1592K& 90.9G & 26.07 & 30.47 \\
IMDN  \cite{hui2019lightweight}  & C &  \textcolor{red}{715K} &  \textcolor{red}{40.9G}& 26.04 & 30.45 \\
SwinIR-light \cite{liang2021swinir} & T  &  930K & 63.6G & 26.47 & 30.92 \\
SRFormer-light \cite{zhou2023srformer} &T & 873K  & 62.8G  & 26.67 &  31.17\\
IPG-tiny \cite{tian2024image} &G & 887K &  61.3G & \textcolor{blue}{26.78} & \textcolor{blue}{31.22} \\
Ours & G & \textcolor{blue}{882K} & \textcolor{blue}{60.1G} & \textcolor{red}{26.89}  & \textcolor{red}{31.33}\\

\bottomrule
\end{tabular}

\label{tab:2}  
}
\vspace{-4mm}
\end{table}

\textbf{Training details.} 
The model processes cropped patches of size $64 \times 64$. To enhance the training dataset, we apply data augmentation techniques such as random flipping and rotations at angles of $[0, 90, 180, 270]$.
Our training spans $500,000$ iterations, using the Adam Optimizer ($\beta_1 = 0.9, \beta_2 = 0.99$) with an initial learning rate of $2^{-4}$. The input size is fixed at $64 \times 64$. A multi-step learning rate strategy is implemented, where the learning rate is halved at specific iterations $[250,000, 400,000, 450,000, 475,000]$. During training, we used $4$ NVIDIA RTX $4090$ GPUs, with a batch size of $8$ set for each card.

\textbf{Training Loss.} 
The standard $L_1$ loss between the super-resolution prediction $I_{SR}$ and the ground truth high-resolution image $I_{HR}$ is utilized for training our models.

\textbf{Evaluation metrics.} 
We assess our model's performance against various lightweight super-resolution (SR) baselines, as presented in Tab.~\ref{tab:1}. The selected baselines include IPG-tiny, a patch-graph-based method, as well as CNN-based models like CRAN \cite{ahn2018fast}, IMDN  \cite{hui2019lightweight}, LAPAR-A  \cite{li2020lapar}, SAFMN \cite{sun2023spatially}  and SeemoRe \cite{zamfir2024see}. Additionally, we feature Transformer-based models such as SwinIR-light \cite{liang2021swinir} , SRFormer-light  \cite{zhou2023srformer}, FIWHN \cite{li2024efficient} and CATANet \cite{liu2025catanet}. We also compared methods using the Mamba framework--MambaIR-light \cite{guo2024mambair}. In particular, we specifically compared with outstanding graph-based methods, such as IPG-tiny \cite{tian2024image} and GMN \cite{gendy2025lightweight}. The performance of the SR models is evaluated at scales of $\times2$, $\times3$, and $\times4$ using PSNR (Peak Signal-to-Noise Ratio) and SSIM(Structural Similarity Index Measure) metrics.

\begin{table}[t]
\caption{Ablation Study on Construct Subgraph Set Block (CSSB) and Subgraph Aggregation Block (SAB).
PSNR is calculated with a scale factor of 4.}
\centering
\begin{tabular}{ccccccc}
\toprule[1pt]
CSSB & SAB & Se5 & Set14 & B100 &  Urban100 & Manga109\\
\midrule
\Checkmark &  \XSolidBrush  &30.87 & 26.72& 24.98 & 24.99  & 30.01\\
 \XSolidBrush &  \Checkmark &31.26 & 27.64& 26.83 & 25.64  & 30.59 \\
\Checkmark &  \Checkmark & \textbf{32.60} & \textbf{28.93}& \textbf{27.78} & \textbf{26.89}  & \textbf{31.33}\\
\bottomrule
\end{tabular}
\label{tab:3}  
\end{table}

\begin{figure*}[ht]
    \centering
    \includegraphics[width=\linewidth]{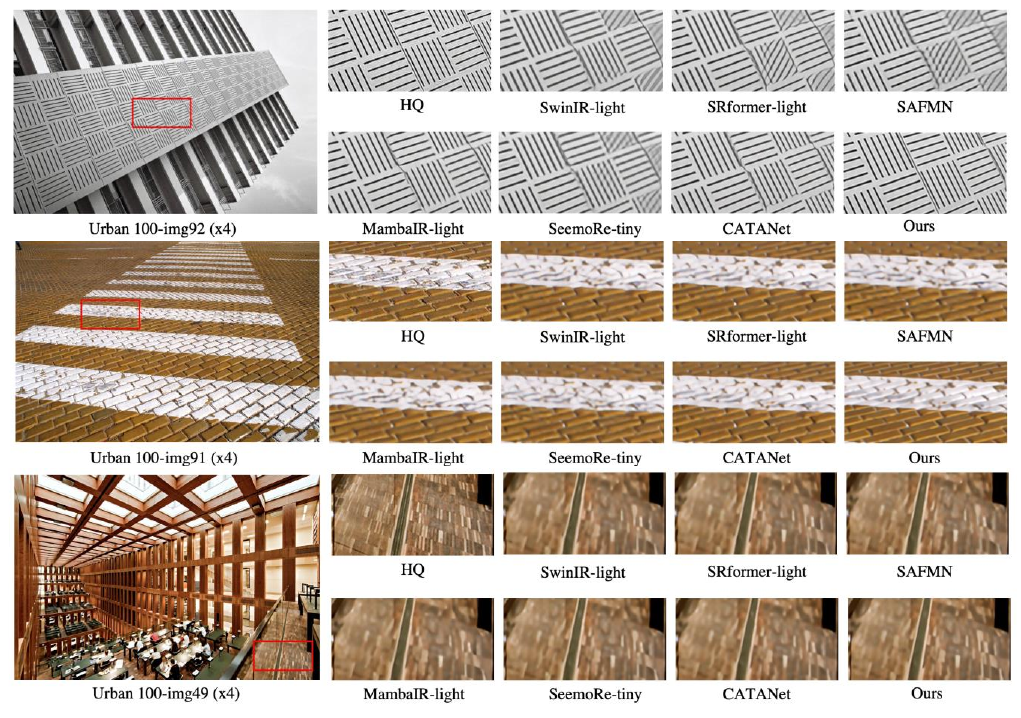}
     \caption{
     Visual results of various methods applied to “img\_92,” “img\_91,” and “img\_49” from the Urban100 dataset at ($\times4$) are presented.
     }
    \label{fig:5}

\end{figure*}

\subsection{Comparisons of Super-resolution Methods}

\subsubsection{Quantitative comparison} 
We conduct a comprehensive quantitative comparison of our method against a range of established approaches, as presented in Tab. \ref{tab:1}. Notably, our method exhibits a significant performance advantage over competing models, indicating its robustness and efficacy in super-resolution tasks. Specifically, HSNet achieves a PSNR of 28.93 dB on the Set14 × 4 benchmark, surpassing the existing state-of-the-art by more than 0.03 dB under standard training conditions, while also demonstrating a notable improvement in SSIM with a score of 0.7886, an increase of 0.0006 compared to the previous best model. This improvement highlights the effectiveness of our method in enhancing image quality.
As illustrated in Tab.~\ref{tab:1}, the proposed method consistently demonstrates remarkable performance in terms of PSNR across various benchmarks.
Furthermore, we conduct a comparison of the parameters and flops of current lightweight SR models on the Urban100 (×4) and Manga109 (×4) datasets. As illustrated in Tab.~\ref{tab:2}, our method not only decreases the model parameters but also reduces computational costs in comparison to earlier Transformer-based and graph-based approaches.

\subsubsection{Qualitative comparison} 
In Fig.\ref{fig:5}, we showcase a series of visual examples that highlight the performance of various super-resolution methods.
By effectively integrating data from multiple subgraphs, HSNet is able to capture a wide range of features at different scales. As shown in Fig.\ref{fig:5}, in Urban100-img92, we can see that our model performs well on the simple stripe restoration task. This multi-scale approach allows the model to address both local textures and global structures, resulting in a comprehensive restoration of images.
Besides, in Urban100-img91, our model demonstrates good performance in handling repetitive textures. 
However, existing methods struggle to satisfactorily restore complex details. For instance, in the case of Urban100-img49, the results indicate that fine textures and intricate structures are often lost. This highlights a key limitation in current approaches, which may rely too heavily on global features while neglecting local detail preservation. Future improvements should focus on enhancing the ability to capture and reconstruct these complex details, possibly by integrating more advanced techniques for local feature extraction and refinement.
In short, our method stands out, showcasing an exceptional ability to accurately restore clean edges while significantly reducing artifacts. This impressive capability is largely due to HSNet's design for capturing precise intricate textures.





\subsection{Ablation Study}

In this section, we carry out ablation studies to thoroughly evaluate and understand each component of the proposed HSNet. To maintain consistency in our comparisons, all experiments were conducted using the ×4 HSNet, with uniform training protocols applied throughout.

\textbf{Effects of the Construct Subgraph Set Block and the Subgraph Aggregation Block.}
We investigate the impact of the Heterogeneous Subgraph Block (HSBlock) on network performance, focusing specifically on its two components: CSSB and SAB. We evaluate the performance by analyzing the PSNR values with a scale factor of 4 across several benchmark datasets: Set14, B100, Urban100, and Manga109, as shown in the Tab.~\ref{tab:3}.
\textbf{CSSB Impact}: The results indicate that incorporating the CSSB significantly enhances PSNR across all datasets. This suggests that the ability of CSSB to extract relevant subgraphs plays a crucial role in improving the quality of the generated images.
\textbf{SAB Impact}: The SAB alone also shows an improvement in PSNR compared to the baseline, indicating its effectiveness in aggregating information from multiple subgraphs. However, its impact is less pronounced than that of the CSSB.
\textbf{Combined Effect}: The model with both CSSB and SAB (HSBlock) consistently outperforms both components individually, demonstrating the synergistic effect of integrating subgraph extraction and aggregation strategies. This improvement highlights the importance of a holistic approach in processing diverse information sources.

\begin{table}[t]
\caption{Ablation Study on Node Sampling Strategy (NSS).
PSNR is calculated with a scale factor of 4.}
\centering
\setlength{\tabcolsep}{3mm}

\begin{tabular}{ccccccc}
\toprule[1pt]
Method & Se5 & Set14 & B100 &  Urban100 & Manga109\\
\midrule
wo NSS &32.42 & 28.71& 27.61 & 26.72 & 31.24\\
w NSS & \textbf{32.60} & \textbf{28.93}& \textbf{27.78} & \textbf{26.89}  & \textbf{31.33}\\
\bottomrule
\end{tabular}
\label{tab:4}  
\end{table}

\textbf{Effects of Node Sampling Strategy.}
We explore the influence of NSS on the performance of the proposed HSNet. NSS is specifically designed to improve the network's capability to capture essential features while simultaneously reducing computational demands. To assess its effectiveness, we compare the performance of HSNet both with and without NSS across several benchmark datasets, including Set14, B100, Urban100, and Manga109.
The PSNR values, calculated with a scale factor of 4, are summarized in Tab.~\ref{tab:4}. 
The results indicate that the integration of the Node Sampling Strategy leads to significant improvements in PSNR across all datasets. For example, in the Set14 dataset, the PSNR increased by 0.22 dB, from 28.71 dB without NSS to 28.93 dB with NSS. 
Those improvements suggest that NSS plays a crucial role in facilitating better feature extraction and representation within the network, ultimately enhancing the quality of image restoration. 

\textbf{Effects of Subgraph Generation Block.}
We explore the influence of the SGB on the performance of the proposed HSNet architecture. 
To evaluate its effectiveness, we compare the performance of HSNet with and without SGB across several benchmark datasets. As shown in Tab.~\ref{tab:5}, the results indicate that the integration of SGB leads to significant improvements in PSNR across all datasets. For example, in the Set14 dataset, the PSNR increased by 1.70 dB, from 27.23 dB without SGB to 28.93 dB with SGB. 
This study underscores the critical importance of SGB in optimizing the performance of HSNet for super-resolution tasks.

\begin{table}[t]
\caption{Ablation Study on Subgraph Generation Block (SGB).
PSNR is calculated with a scale factor of 4.}
\centering
\setlength{\tabcolsep}{3mm}

\begin{tabular}{ccccccc}
\toprule[1pt]
Method & Se5 & Set14 & B100 &  Urban100 & Manga109\\
\midrule
wo SGB &31.74 & 27.23& 26.32 & 25.64 & 30.81\\
w SGB & \textbf{32.60} & \textbf{28.93}& \textbf{27.78} & \textbf{26.89}  & \textbf{31.33}\\

\bottomrule
\end{tabular}

\label{tab:5}  
\end{table}

\begin{table}[t]
\caption{Ablation Study on Graph Aggregation (GA) block.
PSNR is calculated with a scale factor of 4.}
\centering
\setlength{\tabcolsep}{3mm}

\begin{tabular}{ccccccc}
\toprule[1pt]
Method & Se5 & Set14 & B100 &  Urban100 & Manga109\\
\midrule
add &31.23& 27.81& 26.59& 25.71&30.81\\
concat  &32.03& 28.28& 27.14& 26.20&30.72\\
aggregation & \textbf{32.60} & \textbf{28.93}& \textbf{27.78} & \textbf{26.89}  & \textbf{31.33}\\

\bottomrule
\end{tabular}

\label{tab:6}  
\end{table}



\textbf{Effects of Graph Aggregation block.}
We explore the impact of different graph aggregation techniques on the performance of the proposed HSNet, specifically focusing on the integration of subgraphs. We compare three aggregation operations: additive (add), concatenation (concat), and a third method (aggregation). 
The results shown in Tab.~\ref{tab:6} indicate that the graph aggregation method outperforms both the additive and the concatenation methods across all datasets. 
For the various datasets, the PSNR values for GA typically ranged from 26.89 dB to 32.60 dB, while the additive method showed values around 25.71 dB to 31.23 dB, and concatenation ranged from 26.30 dB to 32.03 dB. 
Overall, GA demonstrates an improvement over the other techniques. 
This study emphasizes the importance of selecting the optimal graph aggregation technique, with the GA operation demonstrating superior performance in integrating different subgraphs.

\section{Conclusion}

We present HSNet, a novel Heterogeneous Subgraph Network that integrates information from multiple complementary subgraphs to capture both fine-grained local topologies and broad contextual relationships. At its core lies the Heterogeneous Subgraph Block (HSBlock), which comprises a Construct Subgraph Set Block (CSSB) for generating diverse subgraph views and a Subgraph Aggregation Block (SAB) for fusing these views into enriched feature representations. To balance expressive power with computational efficiency, we introduce a Node Sampling Strategy (NSS) that strategically omits non-critical sampling points while preserving salient information. A concluding Graph Aggregation (GA) block then refines and consolidates these enhanced features, yielding superior overall performance.
Further research on advanced node sampling techniques could improve feature retention and reduce computational complexity. Exploring adaptive sampling methods that dynamically adjust based on graph characteristics could be beneficial.



\normalem
\bibliographystyle{IEEEtran} 
\bibliography{sample}

\end{document}